%% file: Awad-manuscript.tex
\documentclass[manuscript, screen]{acmart}
\usepackage{booktabs} 

\usepackage[ruled]{algorithm2e} 

\usepackage{savesym}
\savesymbol{theoremstyle}
\usepackage{amsmath}
\usepackage{amssymb}
\usepackage{courier}
\usepackage{graphicx}
\usepackage{stmaryrd}
\usepackage{amsfonts}
\usepackage{eucal}
\usepackage{multirow}
\usepackage{subfigure}
\usepackage{framed,color}
\definecolor{shadecolor}{rgb}{0.67,0.67,0.67}

\newcommand{\calA}{\mathcal{A}}

\newcommand{\calL}{\mathcal{L}}

\newcommand{\calP}{\mathcal{P}}

\newcommand{\AF}{\mathcal{AF}}

\newcommand{\Ag}{\mathit{Ag}}

\newcommand{\comp}[1]{\mathit{Comp(#1)}}

\newcommand{\myin}{\mathtt{in}}
\newcommand{\myout}{\mathtt{out}}
\newcommand{\myundec}{\mathtt{undec}}

\newcommand{\calLAP}{\calL\calA\calP}

\acmJournal{TOIT}
\acmVolume{0}
\acmNumber{0}
\acmArticle{0}
\acmYear{2017}
\acmMonth{2}

\acmBadgeR[http://ctuning.org/ae/ppopp2016.html]{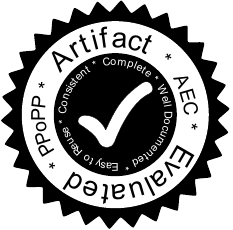}

\setcopyright{acmcopyright}

\acmDOI{0000001.0000001}

\begin{document}
\title{Experimental Assessment of Aggregation Principles in Argumentation-enabled Collective Intelligence} 
\author{Edmond Awad}
\email{awad@mit.edu}
\affiliation{%
  \institution{The Media Lab, Massachusetts Institute of Technology}
  \department{The Media Lab}
  \city{Cambridge}
  \state{MA}
  \postcode{02139}
  \country{USA}
}
\affiliation{%
  \institution{Masdar Institute}
  \department{Electrical Engineering and Computer Science}
  \city{Abu Dhabi}
  \country{UAE}
}
\author{Jean-Fran\c{c}ois Bonnefon}
\affiliation{%
  \institution{Toulouse School of Economics, Center for Research in Management, Institute for Advanced Study in Toulouse, University of Toulouse Capitole}
  \city{Toulouse}
  \country{France}}
  
\author{Martin Caminada}
\affiliation{%
  \institution{School of Computer Science $\&$ Informatics, Cardiff University}
  \country{UK}}

\author{Thomas Malone}
\affiliation{%
  \institution{Sloan School of Management, Massachusetts Institute of Technology}
  \department{Sloan School of Management}
  \city{Cambridge}
  \state{MA}
  \postcode{02139}
  \country{USA}
}
\author{Iyad Rahwan}
\authornote{The corresponding author}
\email{irahwan@mit.edu}
\affiliation{%
  \institution{The Media Lab, Massachusetts Institute of Technology}
  \department{The Media Lab}
  \city{Cambridge}
  \state{MA}
  \postcode{02139}
  \country{USA}
}
\affiliation{%
  \institution{Masdar Institute}
  \department{Electrical Engineering and Computer Science}
  \city{Abu Dhabi}
  \country{UAE}
}

\begin{abstract}
On the Web, there is always a need to aggregate opinions from the crowd (as in posts, social networks, forums, etc.). Different mechanisms have been implemented to capture these opinions such as “Like” in Facebook, “Favorite” in Twitter, thumbs-up/down, flagging, and so on. However, in more contested domains (e.g. Wikipedia, political discussion, and climate change discussion) these mechanisms are not sufficient since they only deal with each issue independently without considering the relationships between different claims. We can view a set of conflicting arguments as a graph in which the nodes represent arguments and the arcs between these nodes represent the defeat relation. A group of people can then collectively evaluate such graphs. To do this, the group must use a rule to aggregate their individual opinions about the entire argument graph. Here, we present the first experimental evaluation of different {principles commonly employed by }aggregation rules presented in the literature. We use randomized controlled experiments to investigate which {principles} people consider better at aggregating opinions under different conditions. Our analysis reveals a number of factors, not captured by traditional formal models, that play an important role in determining the efficacy of aggregation. These results help bring formal models of argumentation closer to real-world application.

\end{abstract}

%
%


%
%



\thanks{Author's addresses: E. Awad, The Media Lab, Massachusetts Institute of Technology, USA; Masdar Institute, UAE, Email: awad@mit.edu; Jean-Fran\c{c}ois Bonnefon, Toulouse School of Economics, Center for Research in Management, Institute for Advanced Study in Toulouse, University of Toulouse Capitole, France, Email: jean-francois.bonnefon@tse-fr.eu; Martin Caminada, School of Computer Science $\&$ Informatics, Cardiff University, UK, Email: CaminadaM@cardiff.ac.uk; Thomas Malone, Sloan School of Management, Massachusetts Institute of Technology, USA, Email: malone@mit.edu; Iyad Rahwan, The Media Lab, Massachusetts Institute of Technology, USA; Masdar Institute, UAE, Email: irahwan@mit.edu.}

\maketitle

\input{samplebody-journals}

\end{document}

%% file: samplebody-journals.tex
\section{Introduction}
In many online systems that support \emph{Collective Intelligence} (e.g. Wikipedia, question answering systems, and discussion forums) different conflicting points of view arise, even based on the same information \cite{apic2011content,marvel2011continuous,introne2011climate}. This raises the challenge of supporting consensus-making among community members. However, many systems on the Web employ arbitrary aggregation rules to handle such tasks, due to the lack of clear appropriate rule given the settings at hand, and the difficulty of evaluating the potential alternatives. Further, the scalability, and the complexity of consensus-making in online systems makes the use of traditional voting rules inadequate.

{
A crucial step towards applying online concensus-making is the representation of information. Argumentation has been shown to provide a realistic environment to represent the conflict on the Web \cite{rahwan2007laying,klein2008supporting,shum:2008}. One of the most influential frameworks is probably the abstract argumentation framework, which was proposed by Dung \cite{dung:1995}. In this framework, arguments are repesented by nodes, and the defeat relations between these arguments are represented by arcs, which form an argumentation graph. This argumentation graph can be evaluated (that is, some arguments are accepted, and others are rejected) in multiple consistent ways. In multi-agent settings, where different agents can subscribe to different evaluations, an aggregation rule is needed to produce a collective consistent evaluation.} Unlike the case with simple aggregation of opinions for isolated propositions, aggregation in the context of argumentation can pose further restrictions to ensure consistency. Consider the following example, which considers making judgments when conflicting information is provided.

\begin{figure}[h!]
\centering
\subfigure[]{\includegraphics[width=10cm]{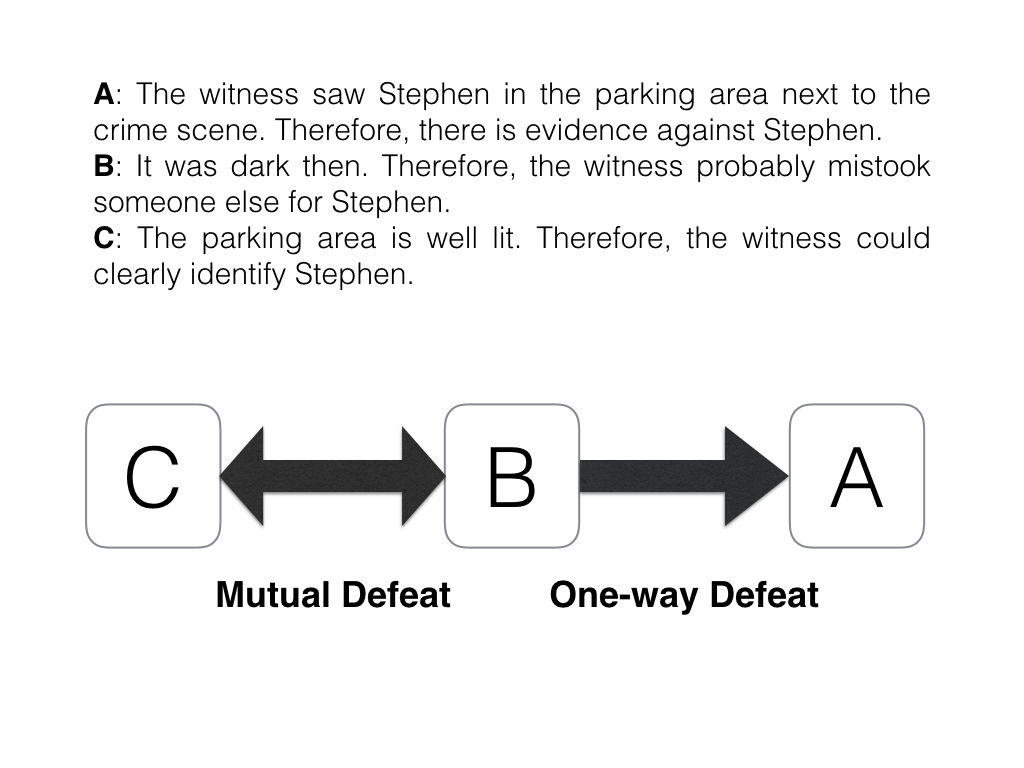}}
\subfigure[]{\includegraphics[width=10cm]{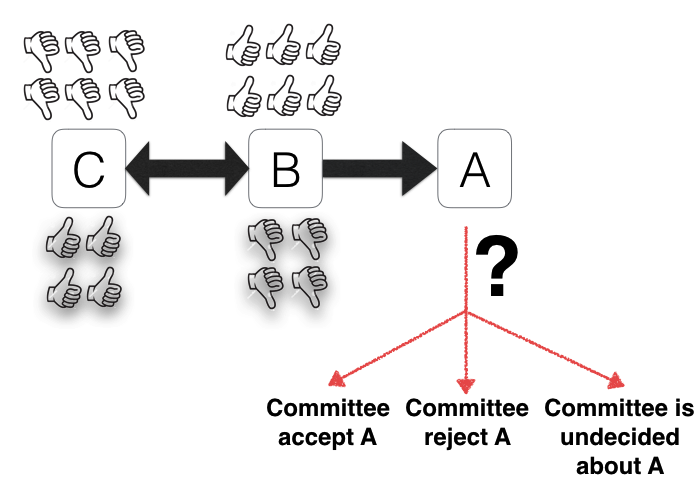}}
\caption{{\bf An example of aggregation in contested domains.} }
\label{fig:Ex1}
\end{figure}

\begin{example}[Suspect Stephen]\label{ex:ext}
\noindent A committee of 10 jury members was formed to make a collective decision about whether there is evidence against Stephen, a potential suspect. The committee is provided with the arguments that were laid down by the opposing sides, to help them make an informed decision. Understandably, arguments of one side are in conflict with the other side's arguments (refer to Figure \ref{fig:Ex1} (a)). The relations among arguments, are represented by an argumentation graph, in which nodes are arguments, and arcs are the defeat relations between arguments (as in Figure \ref{fig:Ex1} (a))). 

Each of the ten members is expected to have a reasonable judgement about which arguments should be accepted. Suppose that six out of the ten members believe that argument $B$ should be accepted, and argument $C$ should be rejected  (thumbs-up/down above graph in Figure \ref{fig:Ex1} (b)), while the other four think that argument $C$ should be accepted and argument $B$ should be rejected (thumbs-up/down below graph). The question is the following: should the committee collectively accept argument $A$, thus accepting there is an evidence against Stephen, or not?

\end{example}

The problem highlighted in the previous example encompasses two issues: First, the provided information is inconsistent. For example, one can see that arguments $B$ and $C$ cannot be accepted together. Likewise, the evaluation of argument $B$ influences that of argument $A$ i.e. if $B$ is accepted, then $A$ should be rejected, and vice versa. As such, the evaluation of argument $A$ is influenced by the evaluation of $B$ and $C$. Aggregating judgments over such a set of inconsistent information is not as simple as it seems. {Since in this example argument $A$ contains the conclusion of interest,} one can suggest two possible methods {as to how collectively evaluate $A$ given indvidiuals' evaluation}: 1) {ignore the individuals' evaluation of arguments $B$ and $C$ and} only aggregate evaluations of argument $A$, and 2) {ignore the individuals' evaluation of argument $A$, and} aggregate evaluation of arguments $B$ and $C$, and then use the outcome to determine the collective evaluation of argument $A$. Applying these two methods on {this example} provides a similar outcome. Unfortunately, these two methods do not always yield the same outcome, as was shown in \cite{awad2015judgment}.\footnote{In fact, this problem mirrors the \emph{discursive dilemma} \cite{pettit2001deliberative} in judgement aggregation (JA). The \emph{discursive dilemma} refers to the paradox that the use of the ``simple'' majority rule to aggregate ``consistent'' individuals' judgments of logically related issues can result into ``inconsistent'' collective judgment of these issues.} Further, choosing between these two methods is not a simple task. There have been many studies discussing the choice between conclusion-based and premise-based methods, the counterparts of the above suggested methods {(1 and 2, respectively)} in other fields of aggregation like judgement aggregation (JA) \cite{bovens2006democratic,pettit2001deliberative,pigozzi2006belief,bonnefon2010behavioral}. This suggests the inadequacy of classical voting in handling such problems.

The second issue is concerned with the amount of support needed to collectively make a judgment, and it mirrors the {issue of choosing among the supermajority rules with the \emph{majority} and the \emph{unanimity} rules being the two extremes. A \emph{supermajority} rule requires that for an alternative to be chosen collectively it has to receive at least $k$ votes, where $k>0.5\times N$ and $N$ is the number of voters ($k=N$ for \emph{unanimity} and $k = \left \lfloor 1+0.5\times N \right \rfloor$ for \emph{majority}).} In the above example, using the majority rule would result in the group concluding that \emph{there is no evidence against Stephen}, while using unanimity would result in the group being undecided about whether \emph{there is evidence against Stephen or not}. Although the comparison between these two methods has been studied extensively, both formally and experimentally \cite{guttman1998unanimity,quesada2011parallel,miller2015group}, it has never been studied experimentally in contested domains where conflicting information are considered in making a collective decision.

Recently, various rules were proposed to appropriately aggregate evaluations of argumentation graphs. The argument-wise plurality rule (AWPR), which chooses the collective evaluation of each argument by plurality was defined and analyzed in \cite{rahwan2010collective,awad2015judgment}. On the other hand, Caminada and Pigozzi \cite{caminada:pigozzi:2010} showed how judgment aggregation concepts can be applied to formal argumentation in a different way. They proposed three possible operators for aggregating labellings, namely the sceptical operator, the credulous operator, and the super credulous operator (collectively shortened here as SSCOs). These operators guarantee not only a well-formed outcome but also a compatible one, that is, it does not go against the judgment of any individual. The analysis of the above methods so far has been restricted to a principle-based approach such as the one devised by Arrow \cite{arrow:1951,arrow:etal:2002}, by evaluating aggregation rules on the base of satisfying seemingly plausible, ``fairness'' postulates. 

In this study, we offer the first experimental-based study of aggregation rules in contested domains. We experimentally compare the desirability of the principles employed by AWPR and SSCOs. We find that principles employed by AWPR are more favorable. However, there can be some conditions in which this is not the case. Our results provide clear suggestions in regard of what aggregation rules to use given some assumed contexts, and offer a first step towards the evaluation of aggregation rules in contested domains.

\section{Theoretical Background: Abstract Argumentation and Labeling Aggregation Rules}
In order to form reasonable judgments on arguments, a formal representation of the relations is needed. Arguments and the relationship between them can be represented as a directed graph in which nodes represent arguments, and arcs between nodes represent binary \emph{defeat} relations over them. This framework is known as abstract argumentation framework, proposed by Dung \cite{dung:1995}. In the previous example, one can see that arguments $B$ and $C$ defeat each other, and argument $B$ defeats argument $A$. 

\begin{figure}[htbp]
    \centering
       \includegraphics[scale=1]{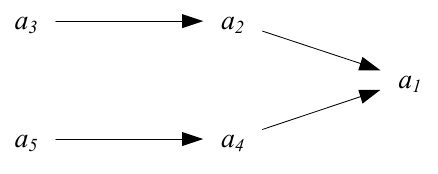}
    \caption{A simple argument graph}
    \label{fig:argument_graph1}
\end{figure}

\begin{definition}[Argumentation framework]
    An \emph{argumentation framework} is a pair $\AF = \langle
    \calA, \rightharpoonup \rangle$ where $\calA$ is a finite set of
    arguments and $\rightharpoonup \subseteq \calA \times \calA$
    is a defeat relation. An argument $a$
    \emph{defeats} an argument $b$ if $(a, b) \in
    \rightharpoonup$ (sometimes written $a ~\rightharpoonup~
    b$).
\end{definition}
For example, in Figure \ref{fig:argument_graph1}, argument $a_1$ is defeated by arguments $a_2$ and $a_4$ which are, in turn, defeated by arguments $a_3$ and $a_5$.

One then can evaluate each argument (i.e. label \cite{caminada:2006,caminada:gabbay:2009} each argument) by accepting it (i.e. labeling it as $\myin$), rejecting it (i.e. labeling it as $\myout$), or being undecided about it (i.e. labeling it as $\myundec$). Formally, a labeling is a total function:

\begin{definition}[Argument labeling]\label{definition:labelling}
Let $\AF = \langle \calA, \rightharpoonup \rangle$ be an argumentation
framework. An \emph{argument labeling} is a total function $L :
\calA \rightarrow \{\myin, \myout,$ $\myundec \}$.
\end{definition}

This evaluation (or labeling), however, should follow some rules. One of the essential {properties}, that is common, is the condition of \emph{completeness}, and is captured, in terms of labelings, in the following two conditions:
\begin{enumerate}
	\item An argument is labeled \emph{accepted} (or \emph{in}) if and only if all its defeaters are rejected (or \emph{out}).
	\item An argument is labeled \emph{rejected} (or \emph{out}) if and only if at least one of its defeaters is accepted (or \emph{in}).
\end{enumerate}
In all other cases, an argument should be labeled \emph{undecided} (or \emph{undec}). Thus, evaluating a set of arguments amounts to labeling each argument using a labeling function $\calL ~:~ \calA \rightarrow \{\myin, \myout, \myundec\}$ to capture these three possible labels. Any labeling that satisfies the above conditions is a \emph{legal labeling}, or a \emph{complete labeling}. These conditions can equivalently be formulated as follows.

\begin{definition}[Complete labeling]\label{definition:CompLabelling}
Let $\AF = \langle \calA, \rightharpoonup \rangle$ be an argumentation
framework. A \emph{complete labeling} is a total function $L :
\calA \rightarrow \{\myin, \myout,$ $\myundec \}$ such that:
\begin{itemize}
    \item $\forall a \in \calA ~:~ \text{if } L(a) = \myin
    \text{ then } \forall b \in \calA ~:~ (b
    \rightharpoonup a ~\Rightarrow~ L(b) = \myout)$; 
    
    \item $\forall a \in \calA ~:~ \text{if } L(a) = \myout
    \text{ then } \exists b \in \calA ~\text{s.t.}~ (b
    \rightharpoonup a ~\wedge~ L(b) = \myin)$; and
    
    \item $\forall a \in \calA ~:~ \text{if } L(a) = \myundec
    \text{ then }$ 
    \begin{itemize}
    \item $\exists b \in \calA ~:~ (b
    \rightharpoonup a \wedge L(b) = \myundec); ~\text{and}~$
		\item $\not \exists b \in \calA ~:~ (b
    \rightharpoonup a \wedge L(b) = \myin)$ 
    
		\end{itemize}
\end{itemize}

\end{definition}

From Example \ref{ex:ext}, one can see that there can be different reasonable positions regarding the evaluation of an argumentation graph (following the previous conditions). Thus, choosing a legal labeling above another becomes a matter of preference. Therefore, in a multi-agent setting, there has been some aggregation rules that were proposed to aggregate preferences over labelings. The work \cite{rahwan2010collective,awad2015judgment} defined and analyzed the argument-wise plurality rule (AWPR) which chooses the collective label of each argument by plurality, independently from other arguments {(i.e. for each argument, among $\myin$, $\myout$, and $\myundec$, the one with the most votes is chosen as the collective label for this argument)}. On the other hand, Caminada and Pigozzi \cite{caminada:pigozzi:2010} proposed three operators for aggregating labelings, namely the sceptical operator, the credulous operator, and the super credulous operator (collectively shortened as SSCOs). {At the crux of these operators is the notion that an argument is not collectively accepted or collectively rejected unless this decision receives a unanimous support. The level of this unanimously required ``support'' though varies across the three rules. While one of the rules defines ``support'' as a strict agreement by everyone to accept (or reject) an argument in order to collectively accept (or reject) this argument, the other two rules are more lenient regarding the support requirement, as they require only some voters to be in a strict agreement to accept (or reject) an argument, in order to collectively accept (or reject) this argument, provided that all other voters are undecided. In details,} in the first of a two-stage procedure, the sceptical operator chooses the label $\myin$ (respectively, $\myout$) as a collective label for an argument when the label $\myin$ (respectively, $\myout$) is chosen by all individuals. Otherwise, an argument is labeled $\myundec$. In the first of the two-stage and three-stage procedures in credulous and super credulous operators, the operator chooses the label $\myin$ (respectively $\myout$) as a collective label for an argument if the label $\myin$ (respectively, $\myout$) is chosen by some individuals, and all other individuals were undecided about this argument. The purpose of the second and third stages of the three operators is to restore consistency in the collective outcome. The three operators guarantee a compatible outcome, that is, the outcome does not go against the judgment of any individual {(refer to the appendix for the formal definitions of the AWPR and SSCOs operators)}. 

The result of using the two types of operators correspond to the two options discussed above. AWPR supports the idea that a label is chosen if it is submitted by the majority regardless what the minority think, while the Sceptical and (Super) Credulous operators (SSCOs) support the idea that minority's opinion should not be completely ignored. 

Analyzing these two types of operators using Arrow's principle-based approach does not provide a clear advantage of one over the other. For example, the following two postulates:

\begin{quote}
{\bf Collective Rationality} \cite{awad2015judgment}: the output of the aggregation should be a \emph{complete} labeling.
\end{quote}

\begin{quote}
{\bf Compatibility} \cite{caminada:pigozzi:2010}: the collective label of every argument does not go against any individual's label of that argument, where $\myin$ and $\myout$ are against each others. 
\end{quote}

are both satisfied by SSCOs but both violated by AWPR. On the other hand, the following two postulates: 

\begin{quote}
{\bf Unanimity} \cite{booth2014interval}: For each argument, every unanimously agreed upon label is chosen as a collective label. 
\end{quote}

\begin{quote}
{\bf Independence} \cite{awad2015judgment}: the collective label of an argument depends only on the votes on that argument.
\end{quote}

are both satisfied by AWPR but both violated by SSCOs. Thus, in the absence of specific preferences over these postulates, other contextual factors may promote the social acceptability of one aggregation rule or the other. In the next section, we consider several such contextual factors and tentatively predict their effects, before reporting a test of these predictions, using \emph{randomized control experiments}, which are the golden standard for doing experimental research.

\section{Contextual Factors}

First, and as explained in  \cite{caminada:pigozzi:2010}, one of the main advantages of SSCOs is that their (compatible) outcomes can be defended by every individual who took part in the decision. {That is, since the collective evaluation chosen by SSCOs is compatbile with each individual's evaluation in the sense that there is no collectively rejected argument that some individuals think should be accepted, and there is no collectively accepted argument that some individuals think should be rejected, then every individual will feel comfortable defending the collective evaluation in public afterwards.} If {laypeople} perceive this advantage, then experimental manipulations that stress the need for everyone involved to defend the outcome should increase preference for {outcomes produced by} SSCOs against {those produced by} AWPR. 

Various such manipulations {(with regards to what the voters are expected to do afterwards)} can be imagined, of which we will test three, from weakest to strongest: a simple reminder of the consequences of the vote; a statement informing participants that each voter is expected to support and defend the group's decision; and a statement informing participants that should the decision of the group prove a mistake, everyone in the group would share responsibility. We deem this last manipulation the strongest because even if group members accept to defend a decision that passed against their vote, they may not be willing to share the blame in case this decision is mistaken. Indeed Ronnegard \cite{ronnegard2015fallacy} argued that the attribution of moral responsibility to all members of a committee is legitimate when the decision is taken through unanimous voting, while it is not necessarily the case otherwise. 

Second, the difference between AWPR and SSCOs {and the principles they use} is mostly about whether to ignore the opinion of the minority or not. Accordingly, people may be more comfortable with AWPR {outcomes} when the minority is very small. To test this hypothesis, we will experimentally manipulate the size of the minority, which will either be small (9 votes to 1) or large (6 votes to 4).

Third, {laypeople} show a preference for more conservative {outcomes} when these outcomes may involve the infliction of personal harm, that is, when an individual will incur a cost or a punishment as a consequence of the decision  \cite{bonnefon2010behavioral}. Once more, because SSCOs are more conservative than AWPR, then decisions that imply to inflict personal harm should shift preferences towards SSCOs {outcomes}. To test this hypothesis, we will present participants with decisions that do or do not imply to inflict personal harm.

\section{Method}
A valid comparison of two aggregation rules would require presenting subjects with the detailed explanation of the aggregation rules, and confirming that subjects fully understand the possible outcomes of each rule given any inputs. This is problematic in user-studies regarding formal entailment (like this study) since general (logical) principles are hard to explain to non-logicians.

An alternative approach is to avoid providing technical details and use examples instead, as used in studies of this form \cite{rahwan2010behavioral,oaksford2004bayesian,stenning1995attitudes}. In our case, that would be presenting subjects with examples, in which voting profiles and their potential outcomes (given each rule) are provided rather than the explanation of the two rules. However, using this approach, in order to make any claim about how two aggregation rules compare with respect to their favorability by people, a systematic comparison of the two rules is required in exhaustive manner by considering all possible types of inputs e.g. different argument graphs, and different labeling profiles. This would naturally increase the number of factors (to account for these variations), and would exponentially increase the number of conditions and number of needed subjects.

As such, in this study, we only consider a subset of these variations. While these variations are not enough to make claims about how the two aggregation rules compare, they are enough to draw conclusions about the favorability of the principles employed by these rules. We will refer to these principles as the \emph{plurality principle}, that is a label that uniquely has the most votes shall be chosen as a collective label, a principle employed by AWPR, and the \emph{compatibility principle}, that is a label is not collectively accepted (resp. rejected) if it is rejected (resp. accepted) by at least one voter, a principle employed by SSCOs.

A total of 400 participants, all US residents, were recruited from the Crowdflower online platform between March 23rd and May 17th, 2015, and were compensated $\$0.25$ each. Each participant read six vignettes that all featured a committee trying to make a collective decision about a conclusion $A$ (see below for an example). Relevant arguments were listed in each vignette, either two in the case of simple argumentation graphs, or four in the case of complex argumentation graphs (this variable did not impact results, and will not be discussed further). The vignettes displayed {the final collective outcome} on each of these arguments, and the task of the participants was to indicate whether, on the basis of these votes, the committee should accept the conclusion $A$ (the AWPR outcome) or declare itself undecided (the SSCOs outcome). Finally, vignettes explained that were the group to declare itself undecided, the decision would be deferred to another body. Full examples are provided in Appendix \ref{sec:vign}.

Of the six vignettes that participants read, three featured a conclusion $A$ that implied to inflict personal harm upon an individual, and three featured a conclusion $A$ that did not imply such a consequence. For example, one vignette featured the conclusion that a football player should be banned for three games (personal harm), whereas one vignette featured the conclusion that the government should stock up on anti-flu drugs (no personal harm).

Participants were randomly assigned to one of eight groups of a 2 $\times$ 4 between-participant design, manipulating the vote ratio and the framing of the decision {(i.e., each participant was assigned to one and only one of the eight groups, and each group was presented with six vignettes. The six vignettes per group share the same vote ratio and the same framing of the decision)}. In one vote ratio condition, all arguments supporting $A$ as well as all arguments rejecting counterarguments to $A$ received 6 votes against 4. In the other vote ratio condition, all arguments supporting $A$ as well as all arguments rejecting counterarguments to $A$ received 9 votes against 1. Finally, the framing of the decision came in four treatments.  In the \emph{Baseline} treatment, no information was provided in addition to the above. In the \emph{Reminder} treatment, one sentence reminded participants of the consequences of accepting $A$ (note that this reminder did not contain any new information). In the \emph{Defense} treatment, one sentence explained that each member of the committee was expected to support and defend the committee's decision. In the \emph{Responsibility} treatment, one sentence explained that if the decision of the committee turned out to be wrong, each member would share the responsibility of this mistake.

\begin{table}[!h]

\caption{An example of a vignette (including a \emph{no personal harm} scenario) shown to participants in the following condition: \emph{Baseline} treatment, and vote ratio is 6:4. The argumentation graph representing the arguments $A$, $B$, and $C$ is similar to the one in Figure \ref{fig:Ex1} (a), but this graph was not shown to participants.}
\footnotesize

\begin{framed}
\begin{flushleft}
\noindent ID: 016

\noindent A governmental committee of 10 members gathered to make a collective decision about whether the government should stock up on medicines against the Mexican flu or not. Consider the following main argument:
\emph{
\begin{itemize}
\item A: The expert virologist says that Mexican flu is a threatening pandemic. Therefore, the government should stock up on medicines against the Mexican flu.
\end{itemize}
}

\noindent In considering argument A, the following need to be taken into account:
\emph{
\begin{itemize}
\item B: This expert has a financial interest in companies making the medicines, according to some journalists. Therefore, his advice cannot be relied upon.
\item C: This expert does not have a financial interest in the companies making the medicines, according to his employer. Therefore, his advice can be relied upon.
\end{itemize}
}

\noindent In the table below, you can see how many members accept (``Yes") or reject (``No") each of the \emph{two} arguments $B$ and $C$.\\

\begin{center}
\footnotesize
   \begin{tabular}{|c|c|c|}
    \hline
     \textbf{No. of Votes} & \textbf{B} & \textbf{C} \\
    \hline
    \hline
    \textbf{6} &  No & Yes \\
        \hline
    \textbf{4} &  Yes & No \\
    \hline
    \end{tabular}
\end{center}
\medskip

Your job is to determine how they will aggregate their votes. Note that we are not asking about your own opinion on the proposed arguments. Rather, we need to know what you think is appropriate for the group to decide given all the provided information above. The options are: 

\begin{enumerate}
\item The group conclude that \emph{the government should stock up on medicines against the Mexican flu}.
\item The group is undecided about whether \emph{the government should stock up on medicines against the Mexican flu or not}. 
\end{enumerate}
In case the committee is undecided, the decision will be postponed for further investigation.\\ 

\end{flushleft}
\end{framed}
\end{table}

\section{Results}

\begin{figure}
\centering
\includegraphics[width=\columnwidth]{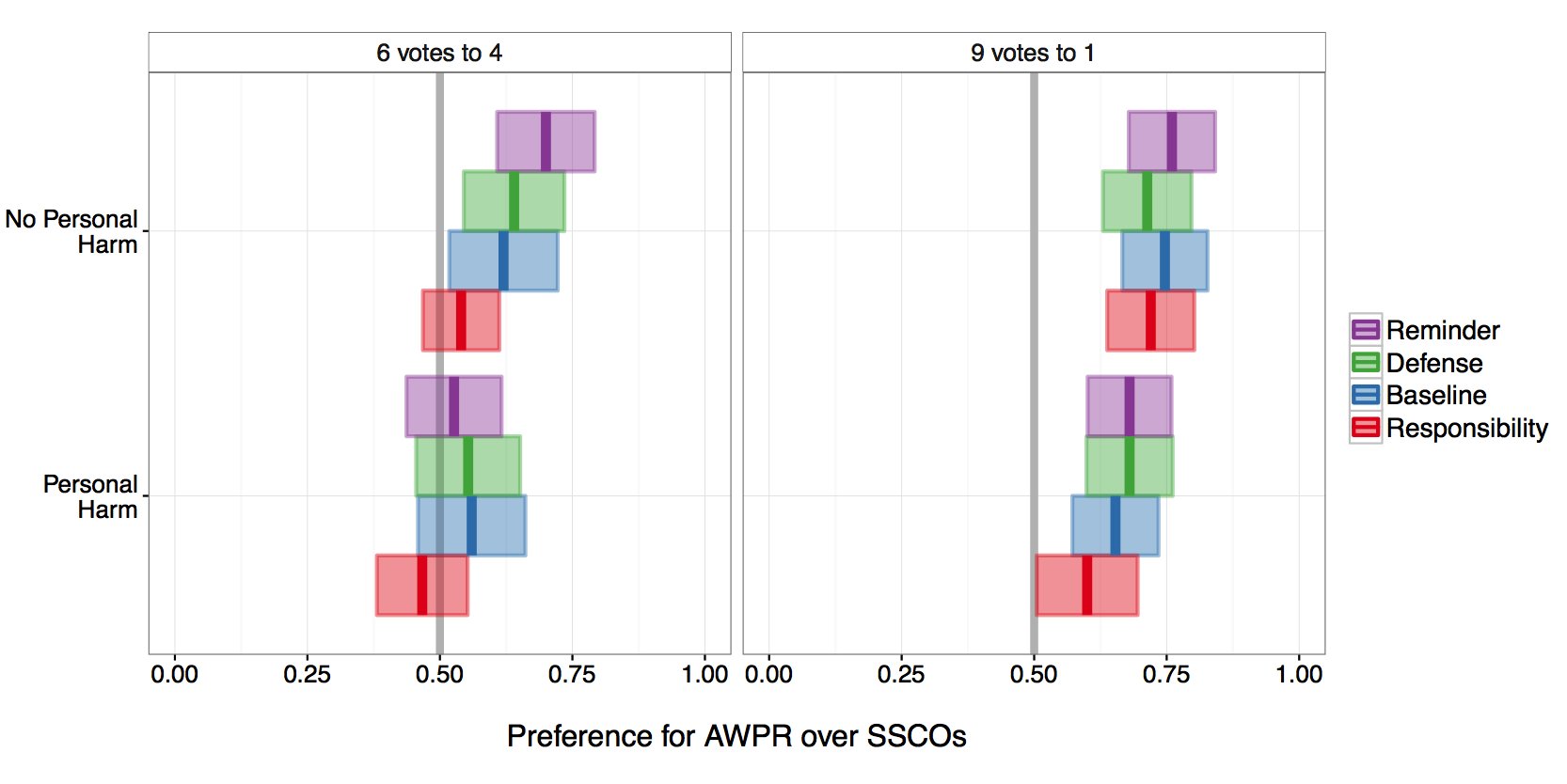}
\caption{Preference for AWPR {outcomes} as a function of decision framing (color-coded), vote ratio, and whether the decision implied to inflict personal harm upon an individual. Boxes shows the average proportion of decisions following AWPR, as well as the 95\%-confidence interval around this proportion.\label{fig:FullPlot}}
\end{figure}

Figure \ref{fig:FullPlot} displays the average proportion of responses favoring \emph{plurality principle}, as a function of the decision framing (color-coded), the vote ratio, and whether the decision implied to inflict personal harm. The width of each box in Figure~\ref{fig:FullPlot} indicates the 95\%-confidence interval for these proportions. A box that overlaps with the grey line indicates that participants in this condition did not show a significant preference for either {principle}.

As suggested by Figure~\ref{fig:FullPlot}, participants showed an overall preference for AWPR {outcomes}. In total, 64\% of responses were in line with AWPR {outcomes}, and this proportion was significantly greater than chance in 11 out of 16 experimental conditions---SSCOs {outcomes}, on the other hand, were never significantly preferred in any experimental condition.

The preference for AWPR {outcomes}, though, was qualified by contextual factors. The visual exploration of Figure~\ref{fig:FullPlot} suggests three main results. First, it seems that the framing of the decision did not significantly affect participants, since the colored boxes are more or less aligned with each other within the four blocks displayed in Figure~\ref{fig:FullPlot}. Second, it appears that participants were less willing to endorse AWPR {outcomes} in the presence of a large minority, since most of the boxes cross the indifference line in the 6 to 4 condition, whereas none of the boxes cross the indifference line in the 9 to 1 condition. Finally,  Figure~\ref{fig:FullPlot} suggests that participants were less willing to endorse AWPR {outcomes} when the decision implied to inflict personal harm, given the leftward shift of the boxes in the personal harm condition.

\begin{table}
\caption{Results of the mixed-design ANOVA analyzing the effect of decision framing, vote ratio, and infliction of personal harm on the preference for AWPR {outcomes}. {The table shows the degree of freedom (DF), the sum of squares (Sum Sq), the mean of squares ($\texttt{Mean Sq} = \frac{\texttt{Sum Sq}}{DF}$), the F test statistics ($F = \frac{\texttt{Mean Sq}}{\texttt{Mean Sq (Residuals)}}$, and the p-value ($P(>F)$). The p-value in the second and the fifth line indicates a significant effect of the vote ratio and infliction of personal harm, respectively.} \label{tab:anova}}
\centering
\begin{tabular}{lrrrrr}
  \hline
 & Df & Sum Sq & Mean Sq & F value & Pr($>$F) \\ 
  \hline
Framing       & 3 & 7.3 & 2.4 & 2.0 & .11 \\ 
  Ratio           & 1 & 25.2 & 25.2 & 20.5 & $<$ .001 \\ 
  Framing:Ratio & 3 & 0.9 & 0.3 & 0.2 & .86 \\ 
  Residuals       & 392 & 481.3 & 1.2 &  &  \\ 
    \hline
  Harm                 & 1 & 14.6 & 14.6 & 25.0 & $<$.001 \\ 
  Framing:Harm       & 3 & 1.1 & 0.4 & 0.6 & .59 \\ 
  Ratio:Harm           & 1 & 0.1 & 0.1 & 0.2 & .64 \\ 
  Framing:Ratio:Harm & 3 & 1.5 & 0.5 & 0.9 & .45 \\ 
  Residuals                   & 392 & 228.6 & 0.6 &  &  \\ 
   \hline
\end{tabular}
\end{table}

These results were confirmed by a mixed-design analysis of variance in which the dependent variable was the proportion of AWPR responses, the between-participant predictors were the vote ratio and the framing condition, and the within-participant predictor was whether the decision implied to inflict personal harm upon an individual. The results of this analysis are displayed in Table~\ref{tab:anova}. As expected, the analysis detected a statistically significant effect of the vote ratio (64\% of AWPR {outcomes} in the 9 to 1 condition, vs. 58\% in the 6 to 4 condition), a statistically significant effect of the infliction of personal harm (59\% of AWPR {outcomes} if personal harm, vs. 68\% if not), but no effect of the framing of the decision. Post-hoc comparisons using a Bonferroni correction indicated that none of the treatments differed significantly from the Baseline treatment (all $p > .31$). {Bonferroni correction \cite{dunn1961multiple} is a multiple-comparison method used to adjust for the incorrect rejection of null hypothesis (Type I error) in the case of several simultaneously tested hypotheses.} Given these results and the statistical power of our analysis (85\% power to detect any effect beyond small, i.e., $f>.25$) we can conclude that even if we had failed to capture a true effect of the framing variable, this effect would likely be small and inconsequential.

\section{Related Work}
The study of aggregation and voting dates back to the 18th century. Since then, different approaches have been employed in order to decide which method is more appropriate to use. The early years of classical voting witnessed an example-based approach adopted by Borda and Condorcet who separately used examples to show the pitfalls of a voting rule, when compared to their own distinct alternative rules \cite{mclean1995classics}. Years later, Kenneth Arrow \cite{arrow:1951,arrow:etal:2002} adopted a more systematic, principle-based approach by evaluating aggregation rules on the base of satisfying seemingly plausible, ``fairness'' postulates. Arrow showed the impossibility of satisfying a small set of these postulates together by any aggregation method. This established the superiority of some rules over others, based on subsets of postulates. Thus, in a given scenario, once the desirable postulates are identified, choosing an appropriate rule becomes a systematic task. However, in reality, identifying the desirable postulates in a specific scenario can be subjective, and dependent on complex factors. Aiming to characterize these complex factors, some studies adopted an experiment-based approach \cite{fiorina1978committee,forsythe1996experimental,guarnaschelli2000experimental,van2010strategic,bassi2015voting}. These studies established the desirability of some aggregation rules given some assumed conditions from a cognitive perspective. 

In the last two decades, aggregation was studied in different new settings including Judgment Aggregation (JA) \cite{list:puppe:2009,list2010theory,list2010introduction,grossi2014judgment}, non-binary JA \cite{dokow2010,dokow2010aggregation}, belief merging \cite{lin1999knowledge}, labeling aggregation \cite{rahwan2010collective,awad2015judgment,caminada:pigozzi:2010,booth2014interval}, and aggregation of {annotated} linguistic resources \cite{endriss2013collective}. While these settings are believed to be closely related, the connection among them and between each of them and the classical voting problem (known as preference aggregation) is not yet fully characterized.

\section{Discussion}

Formal models of reasoning, argumentation, and decision making can be assessed by intuitions, hypothetical examples and simulations---but also by collecting experimental data in the manner of cognitive psychologists, behavioral economists, and experimental philosophers. Indeed, there is a growing interest in the {artificial intelligence} community for assessing the cognitive, psychological plausibility of formal models of reasoning \cite{rahwan2010behavioral,amgoud:etal:2005,Benferhat2005,bonnefon:etal:2008a,dubois:etal:2008}. Here, we {contribute} to this tradition by experimentally identifying the contexts in which human participants would display a preference for SSCOs (aggregation rules that aim at producing compatible outcomes) versus AWPR (the counterpart to the plurality rule in the domain of argumentation). 

Our results {suggest} that {outcomes of AWPR were generally more preferred}, except in situations where (1) the decision implied to inflict personal harm to an individual, and (2) the vote would pass by a narrow margin. The presence of each of these two factors {decreases} the preference for AWPR {outcomes}, and their joint presence {leads} people to hesitate between AWPR and SSCOs {outcomes}. However, and interestingly given previous arguments for using SSCOs, (3) we did not observe an increased preference for SSCOs {outcomes} in situations where all committee members were to defend or take responsibility for the committee's collective decision, independently of their own vote. We now discuss in turn these three findings.

The fact that participants were less likely to endorse {the \emph{plurality principle} employed by} AWPR for decisions that would result in the infliction of harm to an individual is consistent with previous results showing that people prefer more conservative voting rules in such circumstances \cite{Bonnefon2007,bonnefon2010behavioral}. This effect may reflect a concern with avoiding costly, unfair false alarms \cite{List2006}, or the emotional saliency of an identified victim \cite{KogutRitov2005}, but its psychological underpinnings are outside the scope of this article. As a consequence of this finding, though, it may be useful to search for novel voting rules which people deem desirable when they have to deliberate on a personal punishment. Indeed, while we observed that the desirability of AWPR {outcomes} decreased in these circumstances, we did not observe a full switch to a preference for SSCOs {outcomes}.

We also {observe} that participants were less comfortable with {the \emph{plurality principle}} when the vote would pass by a narrow margin (6 to 4 vs. 9 to 1 in our experiment). This result is not surprising: Given that {the \emph{plurality principle}} effectively ignores the opinion of the minority, it makes sense that people are more comfortable with it when the minority that is ignored is small. The applied consequences of this finding are limited, though. Indeed, shifting to SSCOs when the vote is expected to be a close call essentially amounts to deciding in advance that no decision will be reached.

More importantly, we {find} no evidence for one purportedly central advantage of {the \emph{compatibility principle} employed by} SSCOs, that is, their greater desirability when all voters are held accountable (or even responsible) for the committee's decision, independently of their own vote. Participants appeared willing to vote according to AWPR {outcomes} even in these circumstances. One explanation for this finding is that the \emph{plurality principle} is natural enough to appear justified in most situations. Another possibility is that participants weighted against each other the inconvenience for some to defend a decision they opposed, to the inconvenience for all to defend an indecision nobody voted for. Indeed, if $D_o$ is the inconvenience of defending a decision that one opposed, and $D_u$ is the inconvenience of defending an undecided verdict when one was not undecided, then it might be rational to prefer AWPR when $D_u > \frac{m}{N} \times D_o$, where $N$ is the total number of voters and $m$ is the number of voters in the minority. For example, in the case of a 6 to 4 vote, it might still be rational (at least from a utilitarian perspective) to apply AWPR when the inconvenience of defending indecision is at least four tenth of the inconvenience of defending the position that one voted against.

In sum, we {report} the first experimental investigation of the contextual factors that may lead human voters to lean toward {SSCOs} or AWPR aggregation procedures. Although we observed a general preference for {outcomes produced by} AWPR, this preference was moderated by contextual factors. We found null evidence against one important prior claim (SSCOs are good when everyone is expected to defend the collective decision), positive evidence for a vote ratio effect (people like AWPR better when the minority is small), and positive evidence for a moral effect (people like AWPR less when they deliberate about inflicting personal harm to an individual). 

These results provide clear suggestions regarding what aggregation rules are more favorable to use in some contexts, while at the same time they identify contexts in which the favorability of an aggregation rule over another is still open. This opens the door for further work to explore the favorability of new rules in such contexts, on the road towards eventually offering a comprehensive mapping from a given conflicting-domain context to the appropriate aggregation rule to use in that context.

It is important to note that our study compares the two rules using only some variations of argumentation graphs and labeling profiles. Further, the stimuli used probe the favorability of the outcomes of the aggregation rules (rather than the rules themselves). These outcomes can coincide with outcomes by other rules not studied here. As such, the results above are better interpreted on the level of the principles employed by these rules rather than on the level of the rules themselves. In order to make stronger claims regarding the two rules, the consideration of an exhasutive comparison is required, which can be a topic for future work. Future research will {also} identify the psychological mechanisms underlying these preferences, but also the voting rules that people may approve of in situations (e.g., personal harm plus narrow margin) where they seem unsatisfied with AWPR and SSCOs both.

\begin{acks}
Awad is grateful for the fund provided by Masdar Institute to run the experiments. Support through the ANR-Labex IAST is gratefully acknowledged by Bonnefon.
\end{acks}

\appendix
\section*{APPENDIX}
\setcounter{section}{0}

\section{Labeling Aggregation Methods}
For this appendix, we write $\myin(L)$, $\myout(L)$, and $\myundec(L)$ for the set of arguments that are labeled $\myin$, $\myout$, and $\myundec$ respectively by labeling $L$. A labeling $L$ can be represented as $L=(\myin(L)$,$\myout(L)$,$\myundec(L))$. Equivalently, we also denote a labeling $L$ as: $L=\{(A,l)|L(A)=l \text{ for all } A \in \calA, l \in \{\myin,\myout,\myundec\}\}$.

The problem of labeling aggregation can be formulated as a set of individuals that collectively decide how an argumentation framework $\AF = \langle \calA, \rightharpoonup \rangle$ must be labelled. 

\begin{definition}[Labeling aggregation problem \cite{awad2015judgment}]
Let $\Ag=\{1,\ldots,n\}$ be a finite non-empty set of agents, and $\AF = \langle \calA, \rightharpoonup \rangle$ be an argumentation framework. A labeling aggregation problem is a pair $\calLAP= \langle \Ag, \AF \rangle$.
\end{definition}

Each individual $i \in \Ag$ has a labeling $L_i$ which expresses the evaluation of $\AF$ by this individual. A labeling profile is an $|\Ag|$-tuple of labelings.

\begin{definition}[Labeling profile \cite{awad2015judgment}]\label{def:pro}
Let $\calLAP= \langle \Ag, \AF \rangle$ be a labeling aggregation problem. We use $\calL=(L_1,\ldots,L_n) \in \mathbf{L}(\AF)^{|\Ag|}$ to denote a labeling profile, where $\mathbf{L}(\AF)$ is the class of labelings of $\AF$. Additionally, we use $\calL(a)$ to denote the labeling profile (i.e. an $|\Ag|$-tuple) of an argument $a \in \calA$ i.e. $\calL(a)=(L_1(a),\ldots,L_n(a))$.
\end{definition}

The aggregation of individuals' labelings can be defined as a partial function.\footnote{We state that the function is partial to allow for cases in which collective judgment may be undefined (e.g. when there is a tie in voting).}

\begin{definition}[Aggregation function \cite{awad2015judgment}]
Let $\calLAP= \langle \Ag, \AF \rangle$ be a labeling aggregation problem. An aggregation function for $\calLAP$ is a function $F: \mathbf{L}(\AF)^{n} \rightarrow \mathbf{L}(\AF)$. 
\end{definition}
For each $a \in \calA$, $[F(\calL)](a)$ denotes the collective label assigned to $a$, if $F$ is defined for $\calL=(L_{1}, \ldots,L_{n})$.

\subsection{The Argument-Wise Plurality Rule}\label{section:awpr}

The {\em Argument-Wise Plurality Rule (AWPR)} $M$ was proposed in \cite{rahwan2010collective,awad2015judgment}. Intuitively, for each argument, it selects the label that appears most frequently in the individual labelings.

\begin{definition}[Argument-Wise Plurality Rule (AWPR) \cite{awad2015judgment}]
Let $\AF =\langle \calA, \rightharpoonup \rangle$ be an argumentation framework. Given any argument $a \in \calA$ and any profile $\calL =(L_{1}, \ldots, L_{n})$, it holds that $[M(\calL)](a)=l_a \in \{\myin,\myout,\myundec\}$ iff 
$$|\{i: L_{i}(a) = l_a\}| >
    \max_{l'_a \neq l_a}|\{i: L_{i}(a) = l'_a \}|$$

\end{definition}

Note that $M$ is defined for all profiles that cause no ties, i.e. $M(\calL)$ is defined iff there does not exist any argument $a\in\calA$ for which we have at least two labels $l_a$ and $l'_a$ with $l_a \neq l'_a$ and 
\[|\{i: L_{i}(a) = l_a\}| = |\{i: L_{i}(a) = l'_a\}| = \max_l|\{i: L_{i}(a) = l \}|\] 

\subsection{Sceptical and (Super) Credulous Operators (SSCOs)}

The three aggregation operators, namely the {\em Sceptical}, the {\em Credulous} and the {\em Super Credulous}, were defined in \cite{caminada:pigozzi:2010}. In their work, a labeling profile is represented as a set, since the number of votes for each label is irrelevant (as long as it is not zero) for these operators. For convenience, we will assume here that the labeling profile is a tuple (as in Def. \ref{def:pro}).

Crucial to the definitions of the three operators are the notions of \emph{less or equally committed} and \emph{compatible}. A labeling $L_1$ is said to be \emph{less or equally committed} than another labeling $L_2$ if and only if every argument that is labeled $\myin$ by $L_1$ is also labeled $\myin$ by $L_2$ and every argument that is labeled $\myout$ by $L_1$ is also labeled $\myout$ by $L_2$.

\begin{definition}[Less or equally committed $\sqsubseteq$ \cite{caminada:pigozzi:2010}]\label{def:leq}
Let $L_1$ and $L_2$ be two  labelings of argumentation framework $\AF=\langle \calA, \rightharpoonup \rangle$. $L_1$ is less or equally committed as $L_2$ ($L_1 \sqsubseteq L_2$) iff $\myin(L_1) \subseteq \myin(L_2)$ and $\myout(L_1) \subseteq \myout(L_2)$.
\end{definition}

Two labelings $L_1$ and $L_2$ are said to be \emph{compatible} with each other if and only if for every argument, there is no $\myin-\myout$ conflict between the two. In other words, every argument that is labeled $\myin$ by $L_1$ is not labeled $\myout$ by $L_2$ and every argument that is labeled $\myout$ by $L_1$ is not labeled $\myin$ by $L_2$.

\begin{definition}[Compatible labelings $\approx$ \cite{caminada:pigozzi:2010}]\label{def:compat}
Let $L_1$ and $L_2$ be two labelings of argumentation framework $\AF=\langle \calA, \rightharpoonup \rangle$. We say that $L_1$ is compatible with $L_2$ ($L_1 \approx L_2$) iff $\myin(L_1) \cap \myout(L_2) = \emptyset$ and $\myout(L_1) \cap \myin(L_2)=\emptyset$
\end{definition}

The following two definitions are used in the definition of the operators:

\begin{definition}[Initial operators $\large\sqcap$, $\large\sqcup$ \cite{caminada:pigozzi:2010}]
The sceptical initial $\large\sqcap$ and credulous initial $\large\sqcup$ operators are labeling aggregation operators defined as the following:
\begin{itemize}
\item $\large\sqcap((L_1,\ldots,L_n))=$ $\{(A,\myin)|\forall i \in \Ag: L_i(A)=\myin\}$ $\cup$ $\{(A,\myout)|\forall i \in \Ag: L_i(A)=\myout\}$ $\cup$ $\{(A,\myundec)| \exists i \in \Ag:L_i(A)\neq \myin \wedge \exists j \in \Ag: L_j(A)\neq \myout\}$

\item $\large\sqcup((L_1,\ldots,L_n))=$ $\{(A,\myin)|\exists i \in \Ag: L_i(A)=\myin \wedge \neg \exists j \in \Ag:L_j(A)=\myout\}$ $\cup$ $\{(A,\myout)|\exists i \in \Ag: L_i(A)=\myout \wedge \neg \exists j \in \Ag:L_j(A)=\myin\}$ $\cup$ $\{(A,\myundec)| \forall i \in \Ag: L_i(A)=\myundec \vee (\exists j \in \Ag:L_j(A)=\myin \wedge \exists k \in \Ag:L_k(A)=\myout)\}$ 
\end{itemize}
\end{definition}

\begin{definition}[Down-admissible $\downarrow$ and up-complete $\uparrow$ labelings \cite{caminada:pigozzi:2010}]
Let $L$ be a labeling of argumentation framework $\AF=\langle \calA, \rightharpoonup \rangle$. The down-admissible labeling of $L$, denoted as $L$$\downarrow$, is the biggest element of the set of all admissible labelings that are less or equally committed than $L$. The up-complete labeling of $L$, denoted as $L$$\uparrow$, is the smallest element of the set of all complete labelings that are bigger or equally committed than $L$.
\end{definition}

Where \emph{complete} labelings are labelings that satify the three conditions in Def. \ref{definition:CompLabelling}, \emph{admissible} labelings are labelings that satisfy the first two conditions of Def. \ref{definition:CompLabelling}, and \emph{biggest} and \emph{smallest} are defined as with respect to Def. \ref{def:leq}. Now, we provide the definition of the three operators:\footnote{For a better understanding of the three operators, the reader is encouraged to see the clarifying examples of the three operators in \cite{caminada:pigozzi:2010}.}

\begin{definition}[SSCOs: $so_{\AF}$, $co_{\AF}$ and $sco_{\AF}$ \cite{caminada:pigozzi:2010}]
Given an argumentation framework $\AF$, the sceptical $so_{\AF}$, the credulous $co_{\AF}$ and the super credulous $sco_{\AF}$ operators are labeling aggregation operators defined as the following:
\begin{itemize}
\item $so_{\AF}((L_1,\ldots,L_n)) = (\large\sqcap((L_1,\ldots,L_n)))\downarrow$.
\item $co_{\AF}((L_1,\ldots,L_n)) = (\large\sqcup((L_1,\ldots,L_n)))\downarrow$.
\item $sco_{\AF}((L_1,\ldots,L_n)) = ((\large\sqcup((L_1,\ldots,L_n)))\downarrow)\uparrow$.
\end{itemize}
\end{definition}
{Note that the super credulous operator was introduced since the credulous operator can fail to produce a complete collective labeling. This is not the case for sceptical operator, which always produces complete collective labeling.}
\subsection{Postulates}
Inspired by Arrow's principle-based approach, many postulates were defined in the context of argumentation. Most of these postulates are similar to the ones proposed in Judgment Aggregation. We provide here the formal definition of the four postulates that were mentioned in the introduction. \emph{Collective rationality} requires that any possible outcome of the aggregation rule has to be a complete labeling.

\begin{quote}
{\bf Collective Rationality} \cite{awad2015judgment}
For all profiles $\calL$ such that $F(\calL)$ is defined, $F(\calL) \in \comp{\AF}$.
\end{quote}

where $\comp{\AF}$ is the set of all possible complete labelings for the argumentation framrework $\AF$. \emph{Compatibility} requires that the collective label of each argument is compatible (w.r.t Def. \ref{def:compat}) with the label by every individual for that argument.

\begin{quote}
{\bf  Compatibility} \cite{caminada:pigozzi:2010}
For all $i \in \Ag$ and $a \in \calA$ we have $[F_\AF(\calL)](a) \approx L_i(a)$.
\end{quote} 

\emph{Unanimity} requires that for each argument, every unanimously agreed upon label is chosen as a collective label.

\begin{quote}
{\bf Unanimity} \cite{booth2014interval}
For each $a \in \calA$, if there is some $\mathtt{x} \in \{\myin, \myout, \myundec\}$ such that $L_i(a) = \mathtt{x}$ for all $i \in \Ag$ then $[F_\AF(\calL)](a) = \mathtt{x}$. 
\end{quote}

Finally, \emph{independence} requires that the collective label of an argument depends only on the votes on that argument.

\begin{quote}
{\bf Independence} \cite{awad2015judgment}
For any two profiles $\calL =  (L_1,\ldots,L_n)$, $\calL' = (L'_1, \ldots, L'_n)$ such that $F(\calL)$ and $F(\calL')$ are defined, and for all $a \in \calA$, if $L_i(a) = L'_i(a)$ for all $i \in \Ag$, then $[F(\calL)](a) = [F(\calL')](a)$. 
\end{quote}

\section{Argument Sets}\label{sec:argset}
Following, are the six stories that were used in the experiment. The argumentation structure representing the first three is shown in Figures \ref{fig:simp} and the one representing the other three is shown in Figure \ref{fig:comp}. 

\begin{figure}[htbp]
   \centering
      \includegraphics[scale=0.6]{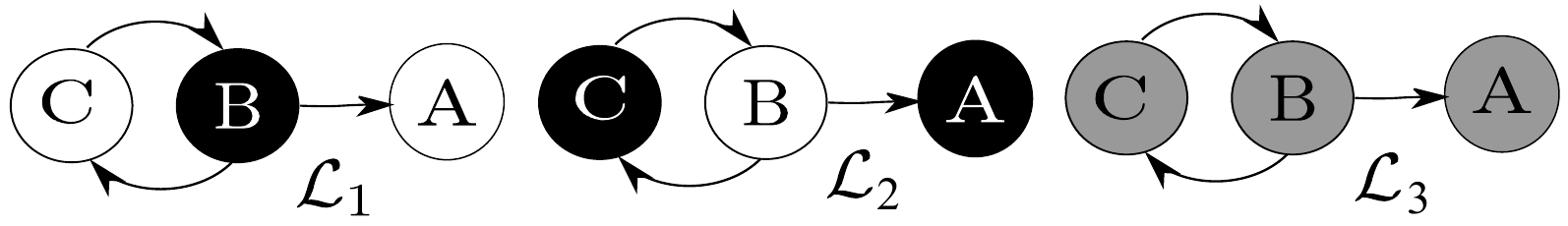}
   \caption{A simple argumentation graph with the three possible \emph{complete labelings}. Nodes are arguments, and arcs are the defeat relations. The color white refers to $\myin$, the color black refers to $\myout$, and the color gray refers to $\myundec$.}
   \label{fig:simp}
\end{figure}

\subsection{Simple AF ({shown in} Figure \ref{fig:simp})}

\begin{enumerate}
\item Argument set 1 - Argument from Knowledge

\underline{Context:} A governmental committee of 10 members gathered to make a collective decision about whether the government should stock up on medicines against the Mexican flu or not.
\begin{itemize}
\item A: The expert virologist says that Mexican flu is a threatening pandemic. Therefore, the government should stock up on medicines against the Mexican flu.
\item B: This expert has a financial interest in companies making the medicines, according to some journalists. Therefore, his advice cannot be relied upon.
\item C: This expert does not have a financial interest in the companies making the medicines, according to his employer. Therefore, his advice can be relied upon.
\end{itemize}
In case the committee is undecided, the decision will be postponed for further investigation.

\item Argument set 2 - Argument from Analogy, Classification and Precedent

\underline{Context:} A school committee of 10 teachers gathered to make a collective decision about whether the school should have a uniform or not.

\begin{itemize}
\item A: After the Central Public School forced students to have a uniform, the attendance of students increased. Therefore, we should have a uniform in our school.
\item B: There can be other factors that contributed to the effect of uniform on attendance. Their case might be different from ours. 
\item C: We share the same entry standards with the Central Public School and our student's families have similar socio-economic status to theirs.\footnote{One might note that, in reality, argument $C$ might not defeat $B$. We only noticed that this can be the case after the experiment was over. However, upon removing this example and redoing the analysis, no major change was found in the results.} 
\end{itemize}
In case the committee is undecided, the decision will be deferred to the school principal who will form another committee.

\item Argument set 3 - Argument from Knowledge

\underline{Context:} A hiring committee of 10 members gathered to make a collective decision about whether a specific candidate is worthy of a good offer or not.
\begin{itemize}
\item A: The candidate's former adviser provided a strong recommendation letter. Therefore, this candidate is worthy of a good offer.
\item B: It is in the adviser's interest that her former student gets a good position. Therefore, she is probably over-selling him.
\item C: The adviser knows she can lose her credibility if the candidate is not as good as she claims. 
\end{itemize}
In case the committee is undecided, the decision will be postponed to get further reference letters.

\end{enumerate}

\begin{figure}[htbp]
   \centering
      \includegraphics[scale=0.6]{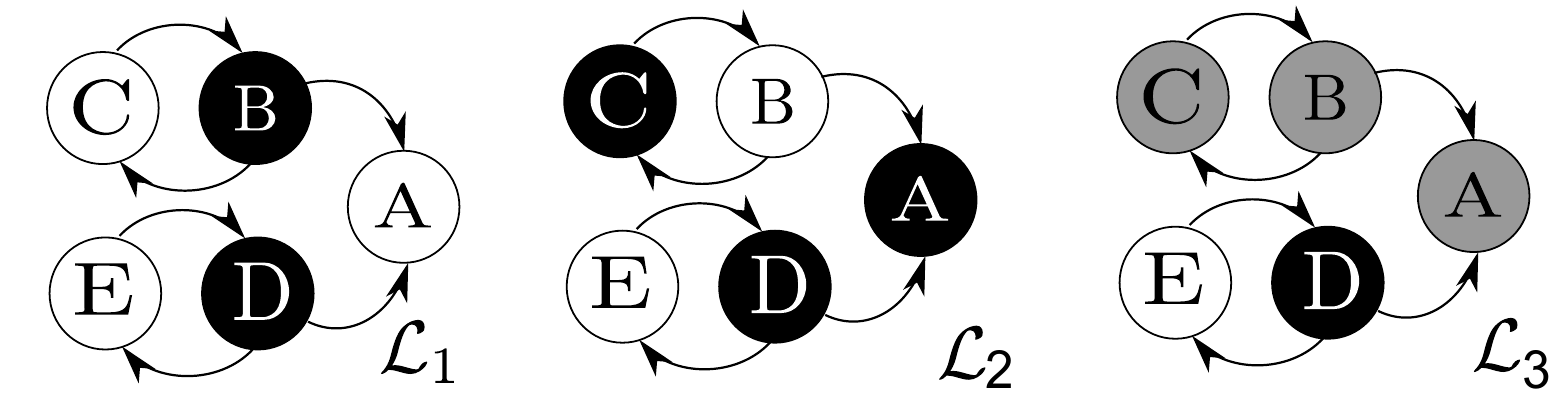}
   \caption{A non-simple argumentation graph with three possible \emph{complete labelings}. Nodes are arguments, and arcs are the defeat relations. The color white refers to $\myin$, the color black refers to $\myout$, and the color gray refers to $\myundec$.}
   \label{fig:comp}
\end{figure}

\subsection{Non-Simple AF ({shown in} Figure \ref{fig:comp})}

\begin{enumerate}
\item Argument set 1 - Argument from Knowledge

\underline{Context:} A committee of 10 jury members gathered to make a collective decision about whether there is evidence against Stephen or not.
\begin{itemize}
\item A: The witness saw Stephen in the parking area next to the crime scene. Therefore, there is evidence against Stephen.
\item B: It was dark then. Therefore, the witness probably mistook someone else for Stephen.
\item C: The parking area is well lit. Therefore, the witness could clearly identify Stephen.
\item D: The witness hates Stephen. Therefore, the witness is biased.
\item E: The witness did not know Stephen well. Therefore, the witness is unbiased.
\end{itemize}
In case the committee is undecided, the decision will be deferred to the judge who will form another committee.

\item Argument set 2 - Argument from Knowledge

\underline{Context:} A group of 10 employees were assigned the task of making a collective decision about whether the company's next summer excursion should be to Niagara Falls or not. 
\begin{itemize}
\item A: The travel agent recommended visiting Niagara Falls. Therefore, the next summer excursion should be to Niagara Falls.
\item B: The travel agent has never been to Niagara Falls. Therefore, we cannot trust his recommendation.
\item C: The travel agent has organized many trips to Niagara Falls before. Therefore, we can trust his recommendation.
\item D: The travel agent recommended visiting a place with natural attractions citing Niagara Falls as an example. He did not specifically recommend Niagara Falls.
\item E: The travel agent specifically recommended visiting Niagara Falls citing its natural attractions as the main reason.
\end{itemize}
In case the committee is undecided, the decision will be deferred to the senior management who will form another committee.

\item Argument set 3 - Argument from Analogy, Classification and Precedent

\underline{Context:} A referees committee of 10 members gathered to make a collective decision about whether the footballer Marconi should be banned for three matches or not. 
\begin{itemize}
\item A: Player Marconi criticized the referee via his official Twitter account. In a previous incident, the footballer Borello was banned for three matches for publicly criticizing the referee. Therefore, Marconi should be banned for three matches.
\item B: Borello criticized the referee in a press conference. That was a different case.
\item C: Both cases are similar in what it matters.
\item D: In another previous incident, the footballer Zotti criticized a referee and got away with it. That was a similar case.
\item E: Zotti's criticism was less direct than the criticism by Marconi and Borello. 
\end{itemize}
In case the committee is undecided, the decision will be postponed for further investigation.

\end{enumerate}

\section{Vignettes}\label{sec:vign}
Here are some concrete examples of vignettes using the argument sets above.

\subsection{Simple AF, Case: ratio $9:1$, \emph{reminder}, Scenario: Uniform}
\begin{table*}[!h]
\footnotesize

\begin{framed}
\begin{flushleft}
\noindent ID: 229

\noindent A school committee of 10 teachers gathered to make a collective decision about whether the school should have a uniform or not. Consider the following main argument:
\emph{
\begin{itemize}
\item A: After the Central Public School forced students to have a uniform, the attendance of students increased. Therefore, we should have a uniform in our school.
\end{itemize}
}

\noindent In considering argument A, the following arguments need to be taken into account:
\emph{
\begin{itemize}
\item B: There can be other factors that contributed to the effect of uniform on attendance. Their case might be different from ours. 
\item C: We share the same entry standards with the Central Public School and our student's families have similar socio-economic status to theirs. 
\end{itemize}
}

\noindent In the table below, you can see how many members accept (``Yes") or reject (``No") each of the \emph{two} arguments $B$ and $C$.\\

\begin{center}
   \begin{tabular}{|c|c|c|}
    \hline
     \textbf{No. of Votes} & \textbf{B} & \textbf{C} \\
    \hline
    \hline
    \textbf{9} &  No & Yes \\
        \hline
    \textbf{1} &  Yes & No \\
    \hline
    \end{tabular}%
\end{center}
\medskip

Your job is to determine how they will aggregate their votes. Note that we are not asking about your own opinion on the proposed arguments. Rather, we need to know what you think is appropriate for the group to decide given all the provided information above. The options are: 

\begin{enumerate}
\item The group conclude that the school should have a uniform. 

\emph{[This will add extra costs on the students' parents' side]}

\item The group is undecided about whether the school should have a uniform or not. 

\emph{[In this case, the decision will be deferred to the school principal who will form another committee]}
\end{enumerate}

\end{flushleft}
\end{framed}
\end{table*}

\newpage

\subsection{Complex AF, Case: ratio $9:1$, \emph{defending}, Scenario: Marconi}

\begin{table*}[h!]
\footnotesize
\begin{framed}
\begin{flushleft}
\noindent ID: 169

\noindent A referees committee of 10 members gathered to make a collective decision about whether the footballer Marconi should be banned for three matches or not. Consider the following main argument:
\emph{
\begin{itemize}
\item A: Player Marconi criticized the referee via his official Twitter account. In a previous incident, the footballer Borello was banned for three matches for publicly criticizing the referee. Therefore, Marconi should be banned for three matches.
\end{itemize}
}
\noindent In considering argument A, the following arguments need to be taken into account:
\emph{
\begin{itemize}
\item B: Borello criticized the referee in a press conference. That was a different case.
\item C: Both cases are similar in what it matters.
\item D: In another previous incident, the footballer Zotti criticized a referee and got away with it. That was a similar case.
\item E: Zotti's criticism was less direct than the criticism by Marconi and Borello. 
\end{itemize}
}

\noindent In the table below, you can see how many members accept (``Yes") or reject (``No") each of the \emph{four} arguments $B$, $C$, $D$, and $E$.\\

\begin{center}
   \begin{tabular}{|c|c|c|c|c|}
    \hline
     \textbf{No. of Votes} & \textbf{B} & \textbf{C}& \textbf{D} & \textbf{E} \\
    \hline
    \hline
    \textbf{9} &  No & Yes & No & Yes\\
    \hline
    \textbf{1} &  Yes & No & Yes & No\\
    \hline
    \end{tabular}%
\end{center}

Your job is to determine how they will aggregate their votes. Note that we are not asking about your own opinion on the proposed arguments. Rather, we need to know what you think is appropriate for the group to decide given all the provided information above. The options are: 

\begin{enumerate}
\item The group conclude that \emph{Marconi should be banned for three matches}.
\item The group is undecided about whether \emph{Marconi should be banned for three matches or not}.
\end{enumerate}
In case the committee is undecided, the decision will be postponed for further investigation.\\

\begin{shaded}
\noindent{\bf\emph{Note that once a collective decision is made, each committee member is expected to support and defend it. It may be awkward for a committee member who disagrees with the group conclusion to defend it to others.}} 
\end{shaded}
\end{flushleft}
\end{framed}
\end{table*}

\subsection{Complex AF, Case: ratio $6:4$, \emph{responsibility}, Scenario: Excursion}

\begin{table*}[h!]
\footnotesize

\begin{framed}
\begin{flushleft}
\noindent ID: 356

\noindent A group of 10 employees were assigned the task of making a collective decision about whether the company's next summer excursion should be to Niagara Falls or not. Consider the following main argument:
\emph{
\begin{itemize}
\item A: The travel agent recommended visiting Niagara Falls. Therefore, the next summer excursion should be to Niagara Falls.
\end{itemize}
}
\noindent In considering argument A, the following arguments need to be taken into account:
\emph{
\begin{itemize}
\item B: The travel agent has never been to Niagara Falls. Therefore, we cannot trust his recommendation.
\item C: The travel agent has organized many trips to Niagara Falls before. Therefore, we can trust his recommendation.
\item D: The travel agent recommended visiting a place with natural attractions citing Niagara Falls as an example. He did not specifically recommend Niagara Falls.
\item E: The travel agent specifically recommended visiting Niagara Falls citing its natural attractions as the main reason.
\end{itemize}
}

\noindent In the table below, you can see how many members accept (``Yes") or reject (``No") each of the \emph{four} arguments $B$, $C$, $D$, and $E$.\\

\begin{center}
   \begin{tabular}{|c|c|c|c|c|}
    \hline
     \textbf{No. of Votes} & \textbf{B} & \textbf{C}& \textbf{D} & \textbf{E} \\
    \hline
    \hline
    \textbf{6} &  No & Yes & No & Yes\\
    \hline
    \textbf{4} &  Yes & No & Yes & No\\
    \hline
    \end{tabular}%
\end{center}

Your job is to determine how they will aggregate their votes. Note that we are not asking about your own opinion on the proposed arguments. Rather, we need to know what you think is appropriate for the group to decide given all the provided information above. The options are: 

\begin{enumerate}
\item The group conclude that the company's next summer excursion should be to Niagara Falls. 

\emph{[If this decision turned out to be wrong, each member would share the responsibility for this mistake]}

\item The group is undecided about whether the company's next summer excursion should be to Niagara Falls or not. 

\emph{[In this case, the decision will be deferred to the senior management who will form another committee]}

\end{enumerate}

\end{flushleft}
\end{framed}
\end{table*}

\bibliographystyle{ACM-Reference-Format}
\bibliography{mybibfile}